\documentclass{ecai} 

\usepackage{latexsym}
\usepackage{amssymb}
\usepackage{amsmath}
\usepackage{amsthm}
\usepackage{booktabs}
\usepackage{enumitem}
\usepackage{graphicx}
\usepackage{color}
\usepackage{algorithm}
\usepackage{algpseudocode}
\usepackage{amssymb}

\newcommand{\BibTeX}{B\kern-.05em{\sc i\kern-.025em b}\kern-.08em\TeX}

\begin{document}


\begin{frontmatter}


\paperid{3983} 


\title{Dynamic Multi-Target Fusion \\for Efficient Audio-Visual Navigation}


\author[A]{\fnms{Yinfeng}~\snm{Yu}
}

\author[A]{\fnms{Hailong}~\snm{Zhang}
\thanks{Corresponding Author. Email: 1160299389@qq.com}
}
\author[B]{\fnms{Meiling}~\snm{Zhu}
}

\address[A]{School of Computer Science and Technology, Xinjiang University}
\address[B]{No. 59 Middle School of Urumqi}


\begin{abstract}

Audiovisual embodied navigation enables robots to locate audio sources by dynamically integrating visual observations from onboard sensors with the auditory signals emitted by the target. The core challenge lies in effectively leveraging multimodal cues to guide navigation. While prior works have explored basic fusion of visual and audio data, they often overlook deeper perceptual context. To address this, we propose the Dynamic Multi-Target Fusion for Efficient Audio-Visual Navigation (DMTF-AVN). Our approach uses a multi-target architecture coupled with a refined Transformer mechanism to filter and selectively fuse cross-modal information. Extensive experiments on the Replica and Matterport3D datasets demonstrate that DMTF-AVN achieves state-of-the-art performance, outperforming existing methods in success rate (SR), path efficiency (SPL), and scene adaptation (SNA). Furthermore, the model exhibits strong scalability and generalizability, paving the way for advanced multimodal fusion strategies in robotic navigation. The code and videos are available at
 https://github.com/zzzmmm-svg/DMTF.
\end{abstract}

\end{frontmatter}


\section{Introduction}

\begin{figure}[!h] 
    \centering 
  \includegraphics[width=0.5\textwidth]{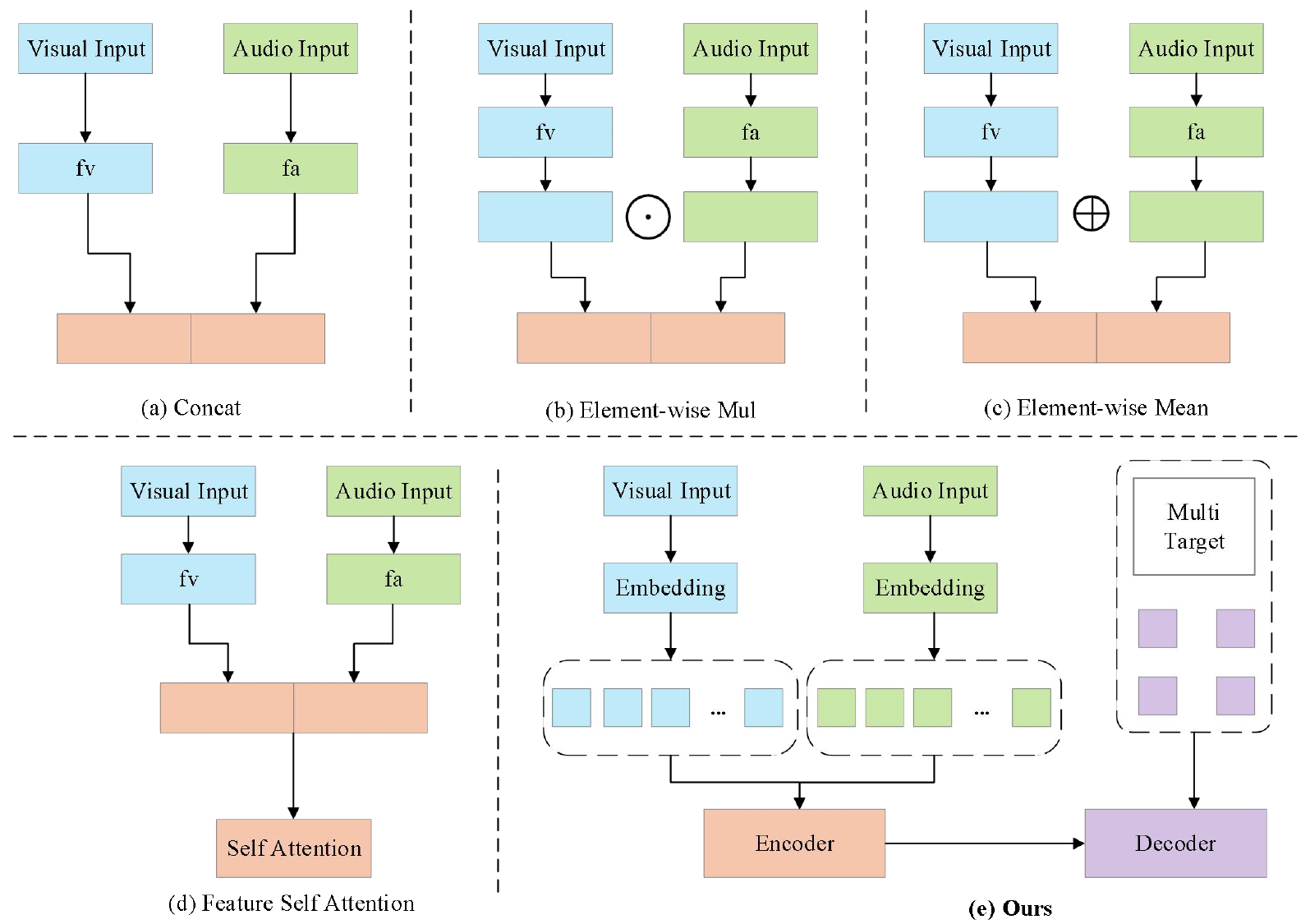}
  \caption{
  A comparison of previous mainstream methods (a-d) with the proposed method (e).
  }
  \label{fig1}
\end{figure}

Embodied intelligence~\cite{zhang2025mapnav,chen2025affordances,chen2024webvln,long2024discuss,wu2024embodied,zheng2024towards} has seen widespread application and in-depth development across various fields, with audio-visual navigation~\cite{chen2020soundspaces,chen2021waypoints,chen2022soundspaces,chen2023omnidirectional} being a critical subdomain within the realm of navigation. Audio-visual navigation~\cite{chen2020soundspaces,chen2021waypoints,chen2022soundspaces,chen2023omnidirectional,ttt, SAAVN, Yu_2022_BMVC, YinfengIJCAI2023MACMA} is irreplaceable in scenarios where intelligent agents operate in environments with limited visibility or poor visual conditions, as they can rely solely on sound to accomplish navigation tasks. From a serious perspective, when an individual experiences a sudden medical emergency, an intelligent agent can promptly identify the individual's location in response to a distress call. In situations such as a gas leak, the agent can detect the alarm sound and swiftly navigate to the designated location to manage the emergency effectively. From a practical, everyday standpoint, when the doorbell rings, the agent can proceed to the door based on the sound of the knock. Similarly, when a dishwasher completes its cycle and emits a notification tone, the agent can move toward the corresponding location in response. These scenarios underscore the critical role of audio-visual navigation and its potential contributions to sustainable development.

To realize such capabilities, innovators have explored various implementations of audio-visual navigation. Nevertheless, challenges persist in improving navigation accuracy. As demonstrated in Figure~\ref{fig1} (a)-(d), previous research has proposed fusion strategies, including concatenation, element-wise multiplication, and addition. While these methods comprehensively utilize audio-visual data, they frequently lack proactivity and selectivity.

To address these challenges, we propose a novel end-to-end framework titled Dynamic Multi-Target Fusion for Efficient Audio-Visual Navigation (DMTF-AVN). In comparison to existing studies, our experimental findings illustrate that this model significantly enhances the interaction and integration between auditory and visual features. This advancement enables agents to expedite the extraction and synthesis of information from multimodal inputs, achieving an efficient synergy between deep learning and reinforcement learning. As illustrated in Figure~\ref{fig1} (e), DMTF maximizes the selective extraction of pertinent information while effectively filtering out redundancy, thus achieving precise and efficient feature extraction and integration. In summary, our primary contributions are as follows:
\begin{itemize}
\item We present an audio-visual navigation methodology that is based on a Dynamic Multi-Target Fusion mechanism, which facilitates selective information processing during the multimodal fusion process. Within this framework, the multi-target module allocates various degrees of freedom to the multimodal data, with each target concentrating on distinct and significant information components. 
\item We conduct comprehensive evaluations of our model on two challenging 3D environment datasets, Replica and Matterport3D, demonstrating superior navigation performance over existing baselines.  
\item Our approach also reveals the dynamic changes in information distribution during the multimodal information selection process. These iterative dynamics facilitate deep and complex interactions between the audio and visual modalities.
\end{itemize}


\section{Related Work}

In audio-visual navigation tasks, the fusion of visual and auditory modalities has a critical impact on navigation performance. Early studies such as SoundNet~\cite{aytar2016soundnet} and Owens et al.~\cite{owens2016ambient} demonstrated the potential of cross-modal alignment for tasks like audio event localization~\cite{lin2019dual,wu2019dual,lin2020audiovisual}. Subsequent work expanded into audio-visual spatialization~\cite{gao20192,chen2020learning,morgado2020learning} and audio-visual parsing~\cite{lin2021exploring,mo2022multi}, fully validating the advantages of multimodal information fusion for understanding and localization. Notably, recent studies~\cite{chen2020soundspaces,chen2021waypoints,chen2022soundspaces} from 2020 to 2022 delve into the challenges and opportunities of integrating visual and auditory signals into end-to-end navigation systems. However, unlike earlier research that mainly focused on capturing complementary features between modalities, our work emphasizes deeper cross-modal interactions and semantic parsing in complex 3D environments to substantially enhance the efficiency and robustness of autonomous agents’ navigation.

End-to-End Object Detection with Transformers~\cite{carion2020end} (DETR), introduced by Facebook AI Research in 2020, represents a groundbreaking advancement in object detection methodologies within the realm of computer vision, leveraging the Transformer architecture~\cite{vaswani2017attention}. This research underscores the substantial benefits that Transformers offer over traditional recurrent neural networks (RNNs) for object detection tasks. DETR reinterprets object detection as a set prediction problem, enabling the model to simultaneously predict the entire array of objects, in contrast to the sequential detection employed by RNNs. The introduction of DETR constitutes a significant milestone in the object detection domain. Drawing inspiration from DETR's innovative approach of utilizing multiple sequences rather than conventional bounding boxes, we have developed a novel multi-target information extraction architecture. This architecture provides a robust theoretical foundation and methodological support for our research, thus considerably enhancing both model design and performance.

\begin{figure*}[htbp] 
    \centering 
    \includegraphics[width=\textwidth,height=0.5\textwidth]{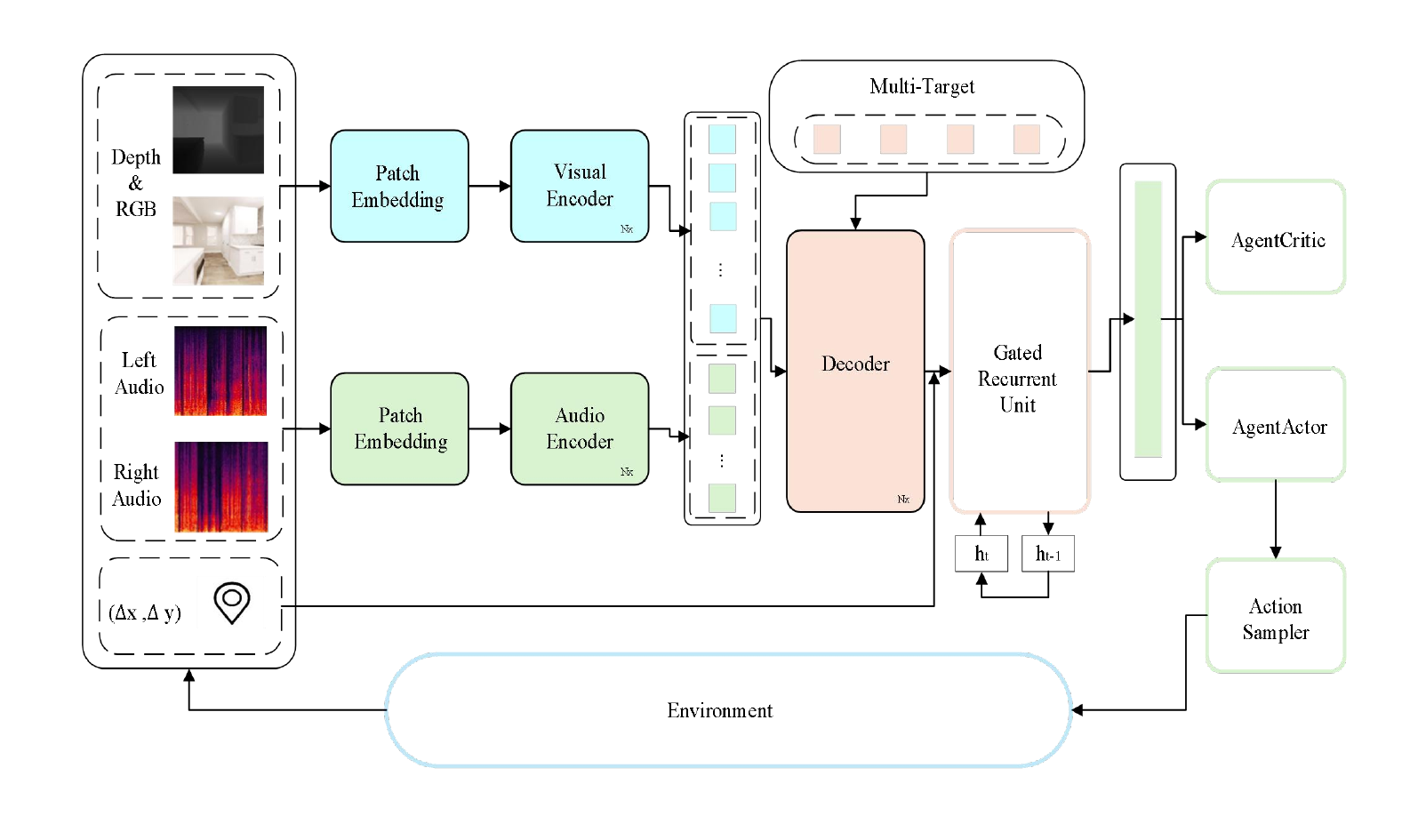} 
    \caption{Audio-Visual Navigation Network. Our model utilizes acoustic and visual cues from the 3D environment to achieve effective navigation through novel feature fusion.} 
    \label{fig2} 
\end{figure*}

Vision Transformer~\cite{DBLP:conf/iclr/DosovitskiyB0WZ21} (ViT) represents a significant advancement in the domain of object detection, with Patch Embedding serving a pivotal function within its architecture. Patch Embedding is tasked with converting raw two-dimensional images into a sequence of one-dimensional patch embedding vectors, thereby facilitating the effective incorporation of Transformer architectures in visual tasks. In comparison to the convolution operations typically utilized in Convolutional Neural Networks (CNNs) or the feature extraction conducted by Multi-Layer Perceptrons (MLPs), Patch Embedding exhibits enhanced adaptability. In our proposed model, Patch Embedding efficiently processes conventional RGB images while also accommodating the preprocessing of diverse modalities, including depth images and Room Impulse Response (RIR) images. This versatility enables the sequence of images transformed by Patch Embedding to be optimally aligned for subsequent feature extraction and multimodal information integration by Transformer models, thereby significantly improving the model's representational capacity and overall performance.


\section{Approach}

In this study, we propose a Dynamic Multi-Target Fusion for Efficient Audio-Visual Navigation (DMTF-AVN), which enables the agent to accurately locate the target sound source through a novel feature fusion method. The model architecture is illustrated in Figure~\ref{fig2}, and it primarily consists of four core modules: (1)Information Extraction: Embedding techniques are used to convert raw modality data into fixed-dimensional embedding vectors. (2) Multi-Target Feature Extraction Strategy: This is the key step to enhancing model performance. The proportional structure between different modality sequences and the number of multi-target sequences significantly influences the quality of feature extraction. A well-designed number of target sequences can effectively extract useful information from multimodal data while filtering out redundant information, thereby optimizing the feature extraction process. (3) GRU~\cite{cho2014learning} (Gated Recurrent Unit): The data sequence processed by the DMTF module is fed into the GRU module, which generates state representations with temporal awareness. GRU is capable of capturing temporal dynamics in the data, providing strong support for the agent’s decision-making process. (4) Actor-Critic Network Architecture: The Actor-Critic framework is employed for action prediction, evaluation, and optimization. The agent makes decisions based on the current state, and the Critic network evaluates these actions based on rewards. The evaluation results are then used to adjust and refine the Actor network’s strategy.  We will introduce each part consecutively in the upcoming paragraphs:

In the Robot Navigation Task, at step \( t \), the robot may receive a specific subset of \( O_t = (I_t, B_t, \Delta) \). For the visual information \( I_t \), we apply the Patch Embedding technique with a patch size of \( 16 \times 16 \) and a stride of 16, converting it into the visual feature vector \( f_I(I_t) \). Processing the auditory information \( B_t \) follows a similar procedure. Still, due to the unique nature of audio data, we first perform concatenation or cropping to ensure the embedded data meets the required dimensions while maintaining the integrity of the information. This results in the auditory feature vector \( f_B(B_t) \). Additionally, the Audio Point Goal task requires an extra input of the relative displacement vector. The relative position vector \( \Delta = (\Delta_x, \Delta_y) \) indicates the displacement from the agent to the target on the 2D ground plane of the scene. Given a visual image or spectrogram \( I \in \mathbb{R}^{H \times W \times C} \), where \( H \) is the height, \( W \) is the width, and \( C \) is the number of channels (e.g. \( C = 3 \) for an RGB image). The image is divided into \( N \) non-overlapping patches, each of size \( P \times P \). The resulting sequence of image patches is denoted as \( f_I(I_t) \)=\( \{I_1, I_2 \dots I_N\} \),\( f_B(B_t) \)=\( \{B_1, B_2 \dots B_N\} \) where 

\begin{equation}
N = \frac{H \times W}{P^2}.
\end{equation}
Multimodal information sequences are combined into an IB information sequence, 
\begin{equation}
\text{IB} = \text{Concat}(f_I(I_t), f_B(B_t)).
\end{equation}

In the Multi-Target Extraction Transformer strategy, multiple blank target sequences are combined with the audio-visual sequences processed by the encoder. These are then processed through a specific multi-head attention mechanism and linear transformation. After decoder block cycles, the resulting information is filtered and integrated, thoroughly combining the visual and auditory multimodal information into a multi-target sequence.
The core formula for Multi-Head Attention can be expressed as:

\begin{equation}
    \text{MultiHead}(\text{MT}_Q, \text{IB}_K, \text{IB}_V) = \text{Concat}(\text{head}_i)W^O,
\end{equation}
where each head is computed as:
\begin{equation}
    \text{head}_i = \text{DMTF-Attention}(\text{MT}_QW_i^Q, \text{IB}_KW_i^K, \text{IB}_VW_i^V),
\end{equation}
Specifically, the attention calculation for each head is:

\begin{equation}
\small
    \text{DMTF-Attention}(\text{MT}_Q, \text{IB}_K, \text{IB}_V) = \text{softmax}\left(\frac{\text{MT}_Q \text{IB}_K^\top}{\sqrt{d_k}}\right){IB}_V.
\end{equation}

\(\text{MT}_Q \in \mathbb{R}^{t_q \times d_k}\) is the Multi-Target Query vector used for information extraction. It represents the content that the model currently aims to query.  
\(\text{IB}_K \in \mathbb{R}^{t_k \times d_k}\) is the Multi-Target Key vector, also used for information extraction, and it is matched with \(Q\).  
\(\text{IB}_V \in \mathbb{R}^{t_k \times d_v}\) is the Multi-Target Value vector, composed of both visual and auditory information. It represents the information sequence to be extracted. Once \(Q\) and \(K\) are successfully matched, the model retrieves the corresponding values from \(V\). \(\sqrt{d_k}\) is a scaling factor used to prevent excessively large dot products that could lead to vanishing gradients. \(W_i^Q \in \mathbb{R}^{d_{\text{m}} \times d_k}\), \(W_i^K \in \mathbb{R}^{d_{\text{m}} \times d_k}\), and \(W_i^V \in \mathbb{R}^{d_{\text{m}} \times d_v}\) are learnable parameter matrices. \(W^O \in \mathbb{R}^{h \cdot d_v \times d_{\text{m}}}\) is the output weight matrix.

DMTF-Attention employs a multi-target mechanism to infer a fixed-size set of predictions via a single pass through the decoder. This strategy dynamically balances cross-modal dependencies in embodied navigation. The core innovation lies in the model's unique scoring method for predicted navigation paths against ground truth paths.

Let \( y \) denote the ground truth set of navigation paths, and \( \hat{y} = \{ \hat{y}_i \}_{i=1}^N \) represent the set of \( N \) predictions, where \( N \) is the number of navigable paths provided by the model for the agent in the scene. If \( N \) exceeds the number of navigation paths in the scene, \( y \) is initialized by padding it with \( \varnothing \) to size \( N \). To establish a bipartite matching between these two sets, we seek a permutation \( \sigma \in \mathfrak{S}_N \) that minimizes the following cost:

\begin{equation}
\hat{\sigma} = \arg \min _{\sigma \in \mathfrak{S}_N} \sum_{i} \mathcal{L}_{\text{match}}\left(y_i, \hat{y}_{\sigma(i)}\right),
\end{equation}
where \( \mathcal{L}_{\text{match}}(y_i, \hat{y}_{\sigma(i)}) \) represents the pairwise matching cost between the ground truth \( y_i \) and the predicted item \( \hat{y}_{\sigma(i)} \).

The matching cost takes into account both the action class prediction and the similarity of the predicted and ground truth paths. Each element \( i \) in the ground truth set can be represented as \( y_i = (c_i, a_i) \), where \( c_i \) is the target action class label (which may be \( \varnothing \)) and \( a_i \in [0, 1]^M \) is a vector used for dynamically selecting important modalities for navigation, with \( M \) being the number of feature vectors. For the predicted item indexed by \( \sigma(i) \), the probability of class \( c_i \) is defined as \( \hat{p}_{\sigma(i)}(c_i) \), and the predicted attention weights are \( \hat{a}_{\sigma(i)} \).

Based on these definitions, \( \mathcal{L}_{\text{match}}(y, \hat{y}) \) is formulated as:

\begin{equation}
\begin{split}
\mathcal{L}_{\text{match}}(y, \hat{y}) = \sum_{i=1}^N \left[ -\mathbf{1}_{\{c_i \neq \varnothing\}} \hat{p}_{\hat{\sigma}(i)}(c_i) + \mathbf{1}_{\{c_i \neq \varnothing\}} \mathcal{L}_{\text{attn}}(a_i, \hat{a}_{\sigma(i)}) \right].
\end{split}
\end{equation}

Here, \( \mathcal{L}_{\text{attn}} \) measures the discrepancy between the ground truth attention weights \( a_i \) and the predicted attention weights \( \hat{a}_{\sigma(i)} \), guiding the model to focus on key modalities for navigation. These parameters are implicitly optimized during the training process using the Actor-Critic algorithm.

A bidirectional GRU equipped with a 512-dim hidden layer is utilized to further convert a sequence of fused embeddings (i.e. \(e_{1}^{o} \ldots e_{T}^{o}\)) into a temporally aware state representation \(s_{t}\). At time \(t\), the GRU cell receives the current embedding \(e_{t}^{o}\) and the previous cell state \(h_{t-1}\) to generate \(s_{t}\) and \(h_{t}\). In essence, \(s_{t} = GRU(e_{t}^{o}, h_{t-1})\).

The state vectors (i.e. \(s_{1} \ldots s_{T}\)) are input into an actor-critic network to (1) predict the conditional action probability distribution \(\pi_{\theta_{1}}(a_{t} | s_{t})\), and (2) estimate the state value \(V_{\theta_{2}}(s_{t})\). The actor and critic are implemented using single linear layers with parameters \(\theta_{1}\) and \(\theta_{2}\), respectively. For simplicity, \(\theta\) is used to denote the combination of \(\theta_{1}\) and \(\theta_{2}\) hereafter. The action sampler in Figure~\ref{fig2} samples the actual action (i.e. \(a_{t}\)) to execute from \(\pi_{\theta_{1}}(a_{t} | s_{t})\). The training objective is to maximize the expected discounted return \(\mathfrak{R}\):

\begin{equation}
\label{eq:6}
\mathfrak{R} = \mathbb{E}_{\pi} \left[ \sum_{t=1}^{T} \gamma^t r(s_{t-1}, a_t) \right].
\end{equation}

Here, \(\gamma\) denotes the discount factor, which is used to balance the importance of future and immediate rewards; \(T\) represents the maximum number of time steps during task execution, indicating the upper bound on the number of actions the agent can take within an episode; and \(\pi\) refers to the policy of the robot agent, which determines the probability distribution over actions based on the current observed state. The reward function \(r(s_{t-1}, a_t)\) describes the feedback provided by the environment at time step \(t\), after the agent performs action \(a_t\) from state \(s_{t-1}\), and serves to guide policy optimization. To improve the stability and convergence efficiency of the learning process, our work adopts the Proximal Policy Optimization (PPO) algorithm~\cite{DBLP:journals/corr/SchulmanWDRK17}, which restricts the update magnitude between the new and old policies to prevent performance collapse, thereby achieving a favorable trade-off between performance and stability. The entire policy optimization and training process is illustrated in Algorithm~\ref{algorithm1}, which includes key steps such as policy sampling, advantage estimation, loss computation, and gradient updates.


\begin{algorithm}
\caption{DMTF-AVN: Multi-Target Information Extraction Transformer
for Audio-Visual Navigation}
\label{algorithm1}
\begin{algorithmic}[1]
\Require
    Environment $\mathcal{E}$, stochastic policies $\pi$, initial actor-critic weights $\boldsymbol{\theta}_0$, initial DMTF weights $\mathbf{W}_0$, \# updates $M$, \# episode $N$, max time steps $T$.
\Ensure
    Trained weights: $\boldsymbol{\theta}_M$ and $\mathbf{W}_M$
\For{$i = 1, 2 \dots M$}
    \State Run policy $\pi_{\boldsymbol{\theta}_{i-1}}$ in environment for $N$ episodes $T$ time steps;
    \State $\{(o_t, h_{t-1}, a_t, r_t)_{i}\}_{t=1}^T \gets \text{roll}(\mathcal{E}, \pi_{\boldsymbol{\theta}_{i-1}}, T)$ at $i$-th update;
    \State Compute advantage estimates;
    \State \textbf{// Optimize w.r.t. $\boldsymbol{\theta}$ and $\mathbf{W}$};
    \State $\boldsymbol{\theta}_i, \mathbf{W}_i \gets$ new $\boldsymbol{\theta}$ and $\mathbf{W}$ from PPO algorithm w.r.t. maximizing Equation (\ref{eq:6});
\EndFor
\end{algorithmic}
\end{algorithm}

\begin{table*}[htbp]
\centering
\caption{Comparison of audio-visual navigation results on the Replica and Matterport3D datasets.}
\label{tab:tab1}
\begin{tabular}{@{}l|cccccc|cccccc@{}}
\toprule
                                         & \multicolumn{6}{c|}{Replica}                                                                                           & \multicolumn{6}{c}{Matterport3D}                                                                                       \\ \cmidrule(l){2-13} 
\multicolumn{1}{c|}{Method}              & \multicolumn{3}{c|}{Heard}                                           & \multicolumn{3}{c|}{Unheard}                    & \multicolumn{3}{c|}{Heard}                                           & \multicolumn{3}{c}{Unheard}                     \\ \cmidrule(l){2-13} 
                                         & SNA $\uparrow$ & SR $\uparrow$ & \multicolumn{1}{c|}{SPL $\uparrow$} & SNA $\uparrow$ & SR $\uparrow$ & SPL $\uparrow$ & SNA $\uparrow$ & SR $\uparrow$ & \multicolumn{1}{c|}{SPL $\uparrow$} & SNA $\uparrow$ & SR $\uparrow$ & SPL $\uparrow$ \\ \midrule
Random Agent~\cite{chen2021waypoints}         & 1.8            & 18.5          & \multicolumn{1}{c|}{4.9}            & 1.8            & 18.5          & 4.9            & 0.8            & 9.1           & \multicolumn{1}{c|}{2.1}            & 0.8            & 9.1           & 2.1            \\
Direction Follower~\cite{chen2021waypoints}   & 41.1           & 72.0          & \multicolumn{1}{c|}{54.7}           & 8.4            & 17.2          & 11.1           & 23.8           & 41.2          & \multicolumn{1}{c|}{32.3}           & 10.7           & 18.0          & 13.9           \\
Frontier Waypoints~\cite{chen2021waypoints}   & 35.2           & 63.9          & \multicolumn{1}{c|}{44.0}           & 5.1            & 14.8          & 6.5            & 22.2           & 42.8          & \multicolumn{1}{c|}{30.6}           & 8.1            & 16.4          & 10.9           \\
Supervised Waypoints~\cite{chen2021waypoints} & 48.5           & 88.1          & \multicolumn{1}{c|}{59.1}           & 10.1           & 43.1          & 14.1           & 16.2           & 36.2          & \multicolumn{1}{c|}{21.0}           & 2.9            & 8.8           & 4.1            \\
Gan et al.~\cite{gan2020look}           & 47.9           & 83.1          & \multicolumn{1}{c|}{57.6}           & 5.7            & 15.7          & 7.5            & 17.1           & 37.9          & \multicolumn{1}{c|}{22.8}           & 3.6            & 10.2          & 5.0            \\
AV-Nav~\cite{chen2020soundspaces}               & 52.7           & 94.5          & \multicolumn{1}{c|}{78.2}           & 16.7           & 50.9          & 34.7           & 32.6           & 71.3          & \multicolumn{1}{c|}{55.1}           & 12.8           & 40.1          & 25.9           \\
AV-WaN~\cite{chen2021waypoints}               & 70.7           & 98.7          & \multicolumn{1}{c|}{86.6}           & 27.1           & 52.8          & 34.7           & 54.8           & 93.6          & \multicolumn{1}{c|}{72.3}           & 30.6           & 56.7          & 40.9           \\
ORAN~\cite{chen2023omnidirectional}                & 70.1           & 96.7          & \multicolumn{1}{c|}{84.2}           & 36.5           & 60.9          & 46.7           & 57.7           & 93.5          & \multicolumn{1}{c|}{73.7}           & 35.3           & 59.4          & 50.8           \\
DMTF (ours)                              & \textbf{76.1}  & \textbf{99.4} & \multicolumn{1}{c|}{\textbf{93.5}}  & \textbf{42.6}  & \textbf{65.8} & \textbf{52.5}  & \textbf{61.9}  & \textbf{95.5} & \multicolumn{1}{c|}{\textbf{81.7}}  & \textbf{38.4}  & \textbf{63.1} & \textbf{54.2}  \\ \bottomrule
\end{tabular}
\end{table*}

\section{Experiment}
\label{sec:experiment}

\subsection{Experimental Setup}  
In this section, we describe the experimental setup used to evaluate the performance of our proposed DMTF architecture for achieving faster audio-visual embodied navigation in 3D environments. Our experiments aim to demonstrate the effectiveness of our approach in integrating audio and visual modalities.

Datasets. We evaluate the proposed model using two datasets: Replica~\cite{straub2019replica} and Matterport3D~\cite{chang2017matterport3d}. Replica~\cite{straub2019replica} and Matterport3D~\cite{chang2017matterport3d} are two high-fidelity 3D scanned datasets widely used in indoor embodied navigation research, encompassing various real-world environments such as residences, offices, and hotels. The spatial coverage of Replica ranges from 9.5 to 141.5 m², while Matterport3D~\cite{chang2017matterport3d} spans a broader scale from 53.1 to 2921.3 m², exhibiting diverse spatial hierarchies and structural complexities. The SoundSpaces framework provides Binaural Room Impulse Responses (BRIRs) for both datasets, with spatial sampling resolutions of 0.5 meters for Replica~\cite{straub2019replica} and 1 meter for Matterport3D~\cite{chang2017matterport3d}. This setup effectively supports a wide range of auditory navigation tasks in both near-field and far-field conditions, offering a robust perceptual foundation and challenging environments for the development of audio-visual navigation systems.

Experimental Environment. The experimental environment is built upon the pioneering work of SoundSpaces~\cite{chen2020soundspaces}, which integrates the open-source platform Habitat~\cite{savva2019habitat} with the two real-world 3D datasets, Replica and Matterport3D, to construct a 3D simulation environment with audio rendering capabilities. This simulated environment enables embodied agents to learn navigation policies, select actions, and move toward target sound sources, even in novel and unmapped environments.


Implementation Details: The model parameters are configured as follows. The RIR sampling rate is set to 44 100 Hz for the Replica dataset and 16 000 Hz for the Matterport3D dataset. The PPO algorithm is trained with 4 epochs, a clip parameter of 0.1, and a value loss coefficient of 0.5. For experiments on the Replica dataset, the learning rate is set to $1 \times 10^{-4}$, while for Matterport3D, it is adjusted to $5 \times 10^{-5}$. Each episode is limited to a maximum of 500 steps during training. The discount factor is set to 0.99. The model is trained for 80,000 agent steps on the Replica dataset and 100,000 agent steps on the Matterport3D dataset.

Evaluation Metrics. We use SR (Success Rate), SPL (Success Weighted by Inverse Path Length), and SNA (Success Weighted by Inverse Number of Actions) as evaluation metrics. SR (Success Rate): 
\begin{equation}
\text{SR} = \frac{1}{N} \sum_{i=1}^{N} S_i,
\end{equation}
the proportion of completed episodes. SPL (Success Weighted by Inverse Path Length): 
\begin{equation}
\text{SPL} = \frac{1}{N} \sum_{i=1}^{N} S_i \frac{l_i}{\max(p_i, l_i)},
\end{equation}
a standard metric that weighs successes based on their adherence to the shortest path. SNA (Success Weighted by Inverse Number of Actions): 
\begin{equation}
\text{SNA} = \frac{1}{N} \sum_{i=1}^{N} S_i \frac{1}{a_i},
\end{equation}
This metric penalizes collisions and in-place rotations by counting the number of actions instead of the path length.

Parameter Explanation: (1) \(N\): The total number of episodes in the test set. (2) \(S_i\): Whether the \(i\)-th episode was successfully completed. If successful, \(S_i = 1\); otherwise, \(S_i = 0\). (3) \(l_i\): The shortest path length of the \(i\)-th episode. (4) \(p_i\): The actual path length of the \(i\)-th episode. (5) \(a_i\): The total number of actions executed in the \(i\)-th episode, including collisions and in-place rotations.

\subsection{Comparison to Prior Work}
In this work, we propose a novel and effective framework for audio-visual embodied navigation in 3D environments. To demonstrate the effectiveness of the proposed DMTF model, we compare it against previous audio-visual embodied navigation baselines~\cite{chen2020soundspaces,chen2023omnidirectional,chen2022soundspaces,gan2020look}.

For the Replica dataset, we report quantitative comparison results in Table~\ref{tab:tab1}. As shown, our method achieves the best results across most metrics under both "Heard" and "Unheard" settings compared to prior audio-visual navigation methods. Specifically, the proposed DMTF outperforms ORAN~\cite{chen2023omnidirectional}, with improvements of 6.0 SNA@Heard \& 2.7 SR@Heard \& 9.3 SPL@Heard, as well as 6.1 SNA@Unheard \& 4.9 SR@Unheard \& 5.8 SPL@Unheard across both settings. Furthermore, when compared to AV-WaN~\cite{chen2021waypoints}, the current state-of-the-art waypoint-based baseline, our model achieves significant performance gains. 

Additionally, Table~\ref{tab:tab1} shows remarkable improvements on the Matterport3D benchmark. Compared to AV-WaN~\cite{chen2021waypoints}, the current state-of-the-art waypoint-based method, we achieve gains of 7.1 SNA@Heard \& 1.9 SR@Heard \& 9.4 SPL@Heard, as well as 7.8 SNA@Unheard \& 6.4 SR@Unheard \& 12.3 SPL@Unheard. Moreover, on the challenging Matterport3D benchmark, our proposed method still outperforms ORAN~\cite{chen2023omnidirectional} by 4.2 SNA@Heard \& 2.0 SR@Heard \& 8.0 SPL@Heard, 3.1 SNA@Unheard \& 3.7 SR@Unheard \& 3.4 SPL@Unheard. We also achieve excellent results compared to AV-Nav~\cite{chen2020soundspaces}, which is a late fusion network based on separate audio and visual encoders. These results highlight the effectiveness of our multi-target strategy for learning fused representations from audio and visual inputs for audio-visual navigation.

\subsection{Comparison of Different Feature Fusion Methods}

\begin{figure}[h] 
    \centering 
    \includegraphics[width=0.5\textwidth]{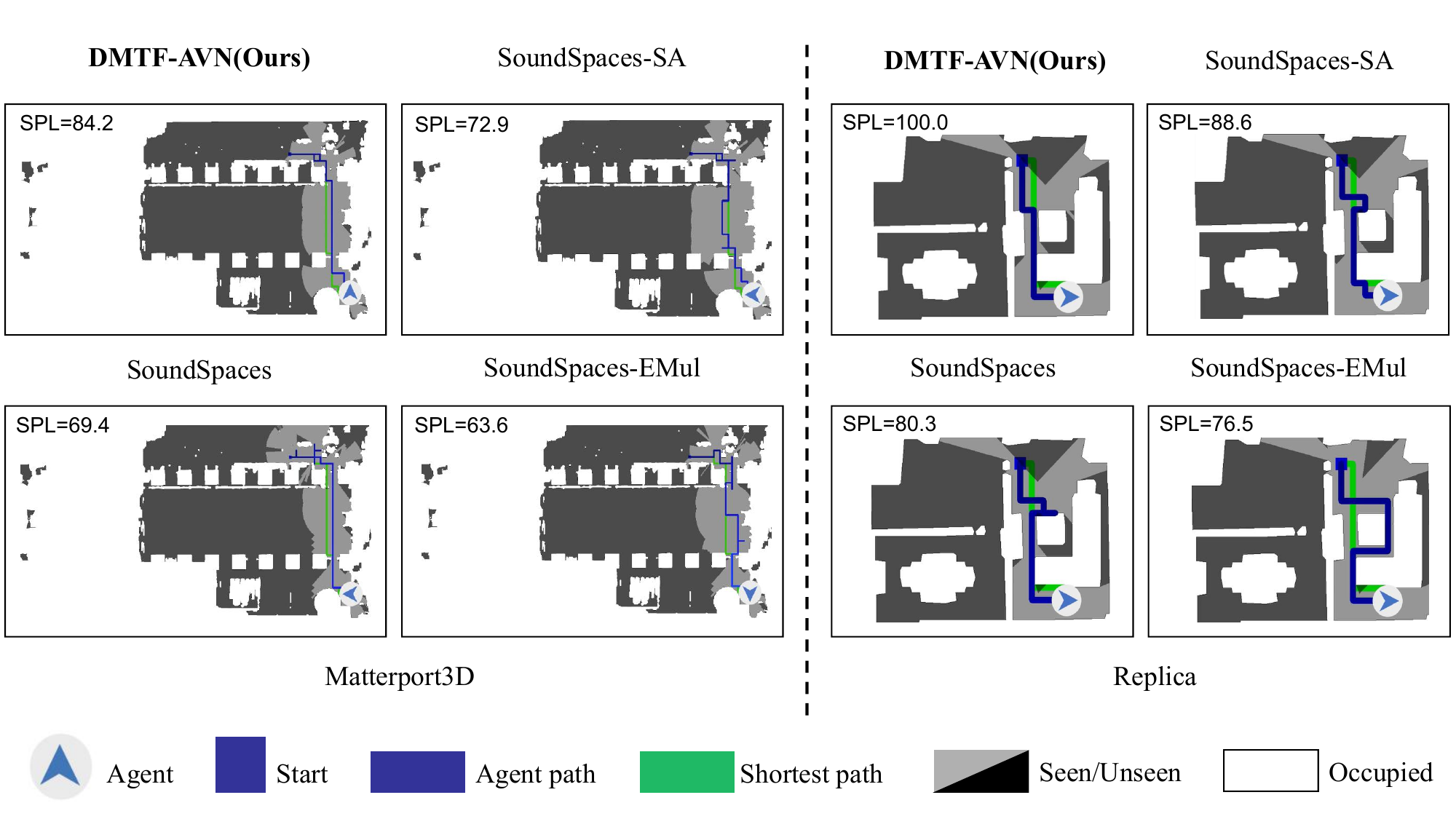}
    \caption{In the Heard condition, the agent’s navigation trajectory has been visualized and recorded on the scene maps of the Replica dataset and the Matterport3D dataset.}
    \label{fig3}
\end{figure}

\begin{table}[]
\centering
\caption{Comparison of Different Feature Fusion Methods.}
\label{tab:tab2}
\resizebox{0.5\textwidth}{!}{
\begin{tabular}{@{}l|l|cc|cc@{}}
\toprule
\textbf{}                   & \textbf{}                   & \multicolumn{2}{c|}{Replica}      & \multicolumn{2}{c}{Matterport3D}  \\ \cmidrule(l){3-6} 
\multicolumn{1}{c|}{Models} & \multicolumn{1}{c|}{Fusion} & Heard           & Unheard         & Heard           & Unheard         \\
                            &                             & SPL($\uparrow$) & SPL($\uparrow$) & SPL($\uparrow$) & SPL($\uparrow$) \\ \midrule
SoundSpaces-EM              & EM                          & 64.8            & 29.9            & 57.0            & 26.3            \\
SoundSpaces-EMul            & EMul                        & 74.6            & 32.2            & 55.9            & 23.5            \\
SoundSpaces                 & Concat                      & 78.2            & 34.7            & 55.1            & 25.9            \\
SoundSpaces-SA              & SelfAttention               & 80.9            & 38.5            & 61.8            & 33.4            \\
DMTF-AVN(Ours)              & DMTF(Ours)                  & \textbf{93.5}   & \textbf{52.5}   & \textbf{81.7}   & \textbf{54.2}   \\ \bottomrule
\end{tabular}
}
\end{table}

We conduct a comprehensive performance evaluation of the proposed DMTF-AVN framework on two representative 3D indoor environment datasets: Replica~\cite{straub2019replica} and Matterport3D~\cite{chang2017matterport3d}. The comparison includes four advanced audio-visual navigation baselines: SoundSpaces-SA, SoundSpaces~\cite{chen2020soundspaces}, SoundSpaces-EMul, and SoundSpaces-EM. Among them, SoundSpaces~\cite{chen2020soundspaces} is widely adopted as a baseline method, employing simple feature concatenation for audio-visual fusion without modeling cross-modal interactions. SoundSpaces-SA enhances the baseline by introducing a self-attention mechanism to better capture inter-feature dependencies. SoundSpaces-EM and SoundSpaces-EMul extend the baseline approach using different fusion strategies—element-wise mean (EM) and element-wise multiplication (EMul) of audio and visual features, respectively—demonstrating various techniques for multimodal fusion. These methods represent a technical progression from simple concatenation to interaction-based modeling in the context of multimodal information integration.

As shown in Figure~\ref{fig3}, the top-down view clearly illustrates that the navigation trajectories generated by DMTF-AVN are more aligned with the shortest paths across multiple test scenes. These trajectories are generally smoother and involve more reasonable turns, highlighting the model’s superior ability to accurately perceive and effectively integrate multimodal information in complex indoor environments. This advantage not only improves navigation efficiency but also significantly reduces unnecessary exploration.

Furthermore, as presented in Table~\ref{tab:tab2}, DMTF-AVN achieves substantial improvements in several key performance metrics, including SR (Success Rate), SPL (Success weighted by inverse Path Length), and SNA (Success weighted by inverse Number of Actions). Among these, SPL and SNA provide a more comprehensive measure of how efficiently an agent accomplishes a task in terms of both trajectory and action efficiency. DMTF-AVN performs especially well in these metrics, underscoring its advantage in learning effective navigation policies.

These experimental results not only systematically validate the effectiveness and superiority of DMTF-AVN in multimodal feature extraction and fusion but also further demonstrate that our proposed multi-target sequence mechanism functions as a set of independent information extraction channels. This mechanism fully leverages the complementarity between different modalities, enhancing the model's ability to perceive and represent critical features. By dynamically focusing on multiple key elements across audio and visual modalities, it boosts performance at both the perception and decision-making levels of navigation.

\subsection{Ablation}
We conducted ablation experiments on the Replica and Matterport3D datasets. Table~\ref{tab:tab3} and Table~\ref{tab:tab4} present the results for various ablation settings.

\begin{table}[]
\centering
\caption{Ablation study on the proposed model on Replica.}
\label{tab:tab3}
\resizebox{\columnwidth}{!}{
\begin{tabular}{@{}c|cccccc@{}}
\toprule
               & \multicolumn{6}{c}{Replica}                                                                                                  \\ \cmidrule(l){2-7} 
Ablation       & \multicolumn{3}{c|}{Heard}                                              & \multicolumn{3}{c}{Unheard}                        \\ \cmidrule(l){2-7} 
\textbf{}      & SNA($\uparrow$) & SR($\uparrow$) & \multicolumn{1}{c|}{SPL($\uparrow$)} & SNA($\uparrow$) & SR($\uparrow$) & SPL($\uparrow$) \\ \midrule
DMTF-AVN(Ours) & \textbf{76.1}   & \textbf{99.4}  & \multicolumn{1}{c|}{\textbf{93.5}}   & \textbf{42.6}   & \textbf{65.8}  & \textbf{52.5}   \\
w/o MTI        & 73.7            & 98.7           & \multicolumn{1}{c|}{89.3}            & 29.8            & 55.2           & 37.5            \\
w/o PE         & 50.0            & 93.2           & \multicolumn{1}{c|}{77.8}            & 21.3            & 52.6           & 34.9            \\
w/o ENSA       & 35.7            & 83.7           & \multicolumn{1}{c|}{65.6}            & 9.1             & 43.8           & 21.7            \\ \bottomrule
\end{tabular}}
\textbf{Note:} “w/o” denotes without.
\end{table}

\begin{table}[]
\centering
\caption{Ablation study on the proposed model on Matterport3D.}
\label{tab:tab4}
\resizebox{\columnwidth}{!}{
\begin{tabular}{@{}c|cccccc@{}}
\toprule
               & \multicolumn{6}{c}{Matterport3D}                                                                                             \\ \cmidrule(l){2-7} 
Models         & \multicolumn{3}{c|}{Heard}                                              & \multicolumn{3}{c}{Unheard}                        \\ \cmidrule(l){2-7} 
\textbf{}      & SNA($\uparrow$) & SR($\uparrow$) & \multicolumn{1}{c|}{SPL($\uparrow$)} & SNA($\uparrow$) & SR($\uparrow$) & SPL($\uparrow$) \\ \midrule
DMTF-AVN(Ours) & \textbf{61.9}   & \textbf{95.5}  & \multicolumn{1}{c|}{\textbf{81.7}}   & \textbf{38.4}   & \textbf{63.1}  & \textbf{54.2}   \\
w/o MTI        & 47.6            & 76.3           & \multicolumn{1}{c|}{65.2}            & 24.9            & 46.2           & 34.4            \\
w/o PE         & 33.5            & 70.2           & \multicolumn{1}{c|}{55.1}            & 12.8            & 39.4           & 25.8            \\
w/o ENSA       & 21.4            & 65.5           & \multicolumn{1}{c|}{46.7}            & 9.3             & 35.8           & 19.1            \\ \bottomrule
\end{tabular}}
\textbf{Note:} “w/o” denotes without.
\end{table}

\subsubsection{w/o PE} Removing the PE module means that audio information is not processed via Patch Embedding, but is instead handled by the auditory CNN encoder. Without PE, the information in the audio sequence fluctuates significantly, which can prevent the multi-target sequences from effectively learning multimodal information during training and ultimately impair prediction accuracy.

\subsubsection{w/o MTI} Removing the MTI module means that the model no longer performs multi-target information extraction, but relies on a single target to extract audio and visual features. MTI uses multiple extraction sequences to focus on the key features of both the visual and audio streams, allowing the model to concentrate on several important aspects of the multimodal information.

\begin{figure}[htbp] 
    \centering 
    \includegraphics[width=0.5\textwidth,height=0.27\textwidth]{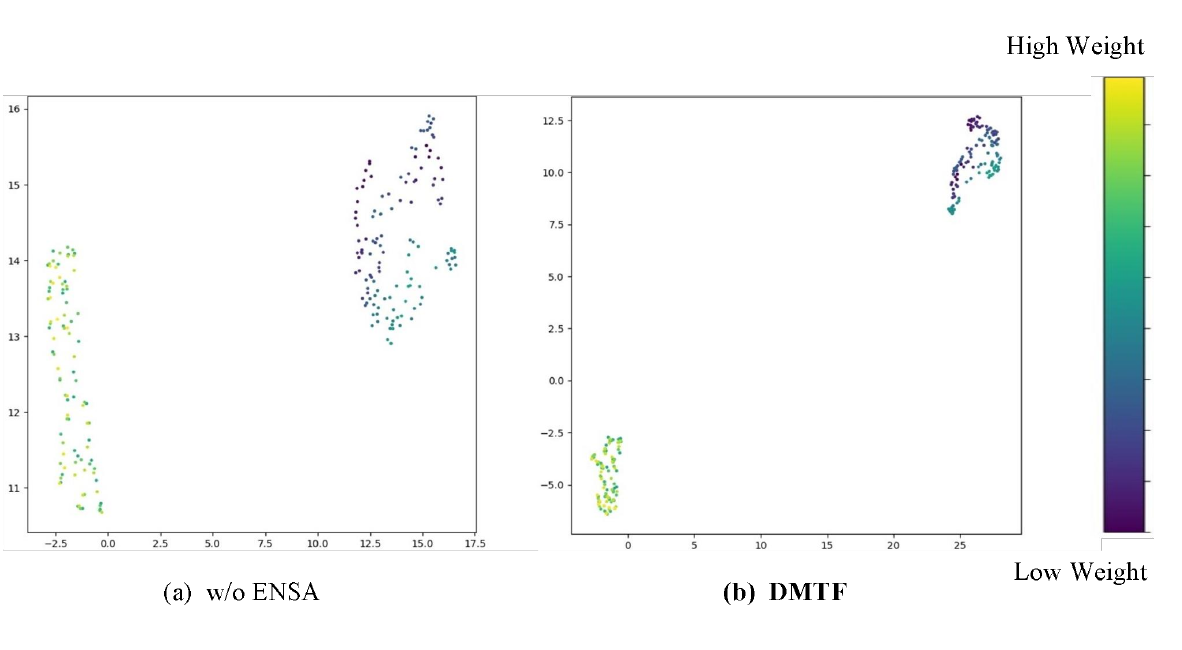}
    \caption{Comparison of attention weights on multimodal information sequences between the w/o ENSA setting and the complete DMTF module.}
    \label{fig4}
\end{figure}

\subsubsection{w/o ENSA} The removal of the ENSA module necessitates the concurrent elimination of both the visual and audio encoders within the model, resulting in a notable decline in overall model performance. The multi-head self-attention encoding layers present in the visual and audio encoders evaluate the similarities between elements in the sequence, subsequently assigning varying weights to them. This process facilitates a nuanced modulation of multimodal information. As illustrated in Figure~\ref{fig4}, the implementation of UMAP~\cite{DBLP:journals/jossw/McInnesHSG18} offers an intuitive visualization, demonstrating that the presence of the ENSA module concentrates attention weights, thereby enhancing the model's ability to extract critical information more efficiently.

These ablation results demonstrate that each module is essential and works synergistically to enhance the model's ability to make accurate path selections.

\subsection{Visualization}

\begin{figure}[h] 
    \centering 
    \includegraphics[width=0.5\textwidth]{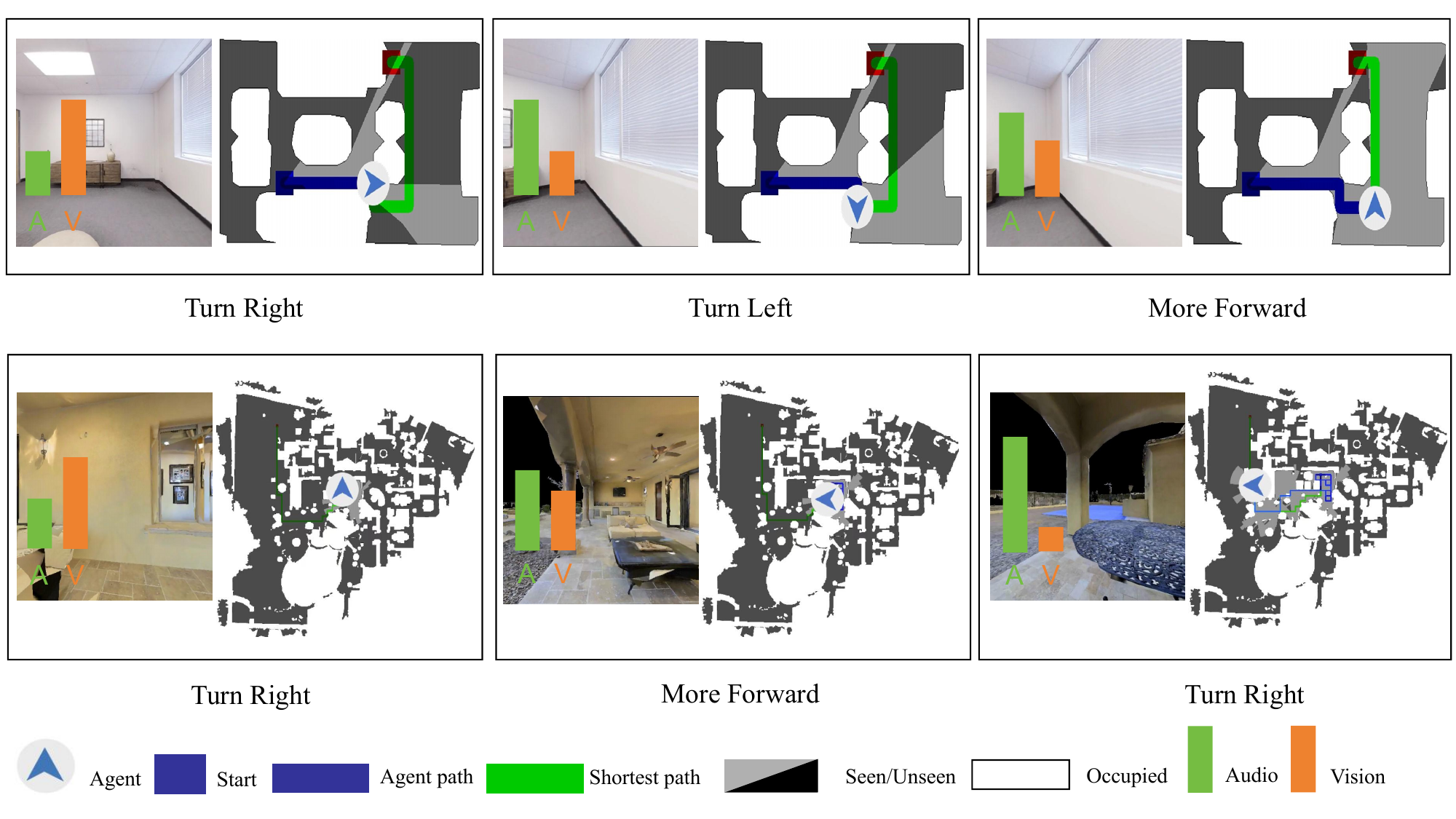}
    \caption{Dynamic visual and echo impact for two episodes. Columns correspond to three sampled time steps. The green and orange bars represent the importance of audio and vision, respectively.}
    \label{fig5}
\end{figure}

\begin{figure}[h] 
    \centering 
    \includegraphics[width=0.5\textwidth]{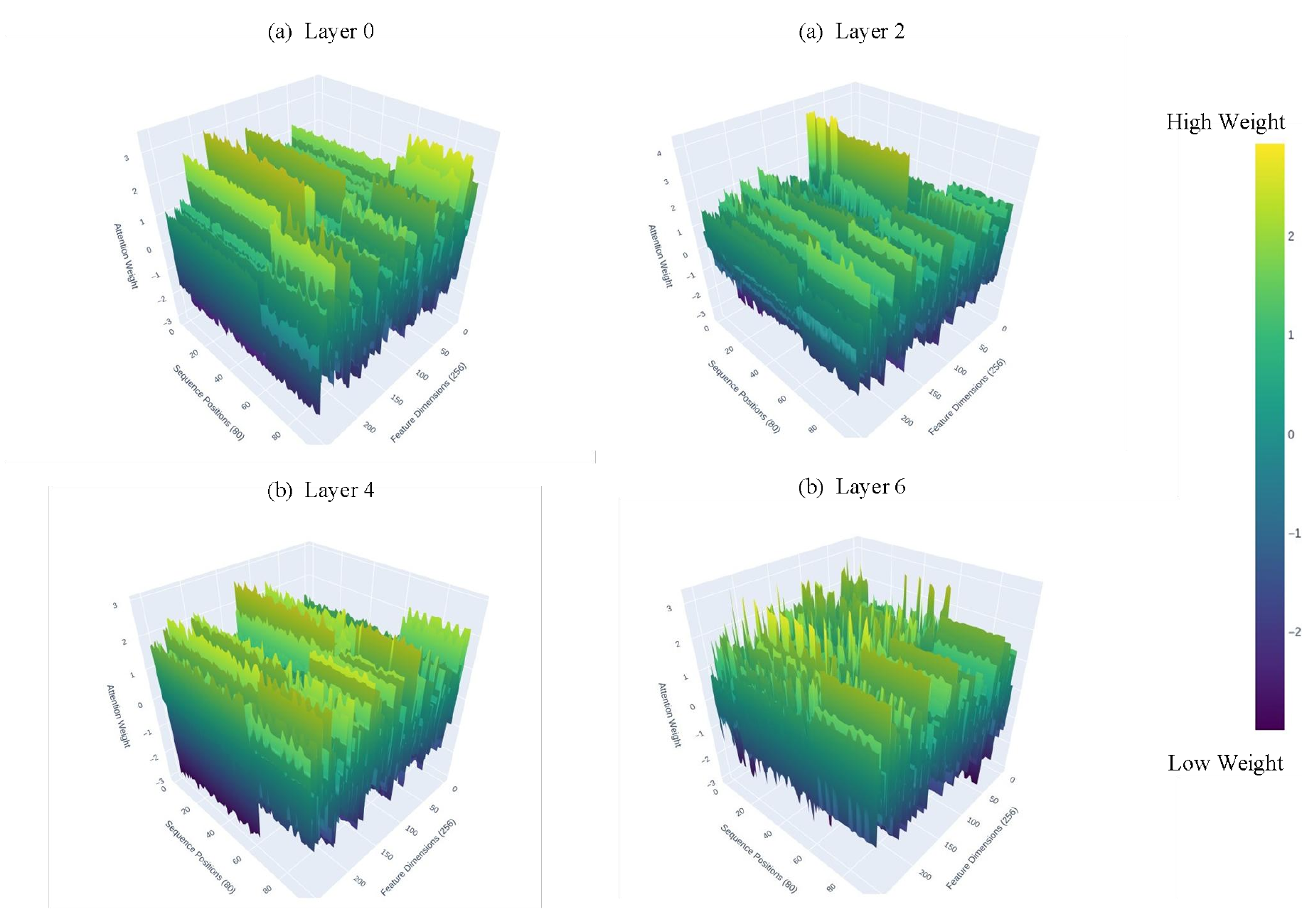}
    \caption{3D Attention Weight Distribution of Visual and Audio Features across Encoder Layers from Replica Dataset.}
    \label{fig6}
\end{figure}

In the robot navigation task, visual and auditory information in the environment dynamically changes over time, resulting in varying impacts of these inputs on the robot's behavior. As shown in Figure~\ref{fig5}, to evaluate the specific influence of visual and auditory information on the robot’s decision-making process, we developed a deep learning-based dynamic weight computation framework. This framework computes dynamic weights for the time steps corresponding to key actions in each episode and normalizes them to quantify the relative contributions of the visual and auditory modalities to the navigation task.

In the DMTF-AVN framework, the visual and auditory information sequences, after being processed by Patch Embedding, are fed into their respective visual and auditory encoders, enabling the initial extraction and representation of modality-specific features. Figure~\ref{fig6} presents a 3D visualization of the attention weight distributions for the multimodal information sequences across different encoder layers. From the figure, it can be observed that as the number of encoder layers increases, the attention gradually shifts from being concentrated on a few local regions to being more widely distributed across various feature locations. This evolution in attention distribution reflects the model’s transition from focusing on local details to achieving a more global and abstract semantic understanding. Such a dispersed attention mechanism enables the model to effectively select and fuse key information from multiple modalities, thereby facilitating the formulation of more accurate navigation strategies. This phenomenon indicates that the DMTF-AVN framework exhibits progressively enhanced perception and reasoning capabilities in multimodal interaction.

\section{Conclusion}

In this work, we present DMTF-AVN, a novel framework for audiovisual embodied navigation that achieves efficient 3D environment exploration through dynamic multimodal fusion. Our method introduces a multi-target mechanism to filter and process heterogeneous sensory data, enabling the agent to adaptively prioritize critical visual-audio features via collaborative interactions between the visual encoder, auditory encoder, and multi-target sequences. Extensive evaluations on the Replica and Matterport3D datasets validate the superiority of DMTF-AVN: Compared to existing multimodal fusion baselines, in the Heard setting on the Replica dataset, it achieves improvements of 6.0\% in success rate (SR), 2.7\% in success weighted by inverse path length (SPL), and 9.3\% in scene navigation adaptability (SNA). These results highlight the effectiveness of multi-target information extraction in complex navigation scenarios.

Despite these advancements, our approach has limitations. First, DMTF-AVN relies on synchronized high-quality audiovisual inputs, which may be challenged by real-world sensor noise or temporal misalignment. Second, while our framework currently focuses on vision and audio modalities, extending it to incorporate complementary sensors (e.g., LiDAR, infrared) could further enhance robustness. Future work will explore hierarchical fusion architectures and self-supervised adaptation strategies to address data quality variations.

This research makes two key contributions: (1) We propose the first transformer-based multi-target fusion mechanism that dynamically balances cross-modal dependencies for embodied navigation, and (2) We empirically demonstrate that deep audiovisual integration with context-aware feature selection substantially improves navigation efficiency in photorealistic 3D environments. By bridging the gap between theoretical multimodal fusion and practical robotic deployment, our work provides a foundational framework for next-generation autonomous agents operating in unstructured real-world settings.




\begin{ack}

This research was financially supported by the National Natural Science Foundation of China (Grant No. 62463029) and the Natural Science Foundation of Xinjiang Uygur Autonomous Region (Grant No. 2015211C288).

\end{ack}


\bibliographystyle{IEEEtran}   
\bibliography{mybib}      

\end{document}